\documentclass[journal]{IEEEtran}
\usepackage{color}

\usepackage{cite}
\usepackage{amsmath,amssymb,amsfonts}
\usepackage{amssymb}
\usepackage{cleveref}
\usepackage[version=3]{mhchem} 
\usepackage{float}
\usepackage{color,soul}               
\usepackage{enumerate}
\usepackage{algorithmic}
\usepackage{graphics}
\usepackage{epsfig}
\usepackage{amssymb}
\usepackage{adjustbox}
\usepackage{url} 
\usepackage{multirow}
\usepackage{subfigure}
\usepackage{booktabs}
\usepackage{multicol}
\usepackage{array}
\usepackage{tabularx}
\usepackage{flushend}
\usepackage{bm}
\usepackage{blindtext}
\usepackage{nicefrac}
\usepackage{textcomp}
\usepackage{pbox}
\usepackage{url}

\begin{document}
%

\title{A Survey on AI Sustainability: Emerging Trends on Learning Algorithms and Research Challenges}

 
%
%
%

\author{Zhenghua Chen, 
        Min Wu, 
        Alvin Chan, 
        Xiaoli Li, 
        and~Yew-Soon Ong,~\IEEEmembership{Fellow,~IEEE}


\thanks{Zhenghua Chen and Min Wu are with Institute for Infocomm Research, A$^*$STAR, Singapore (Email: chen0832@e.ntu.edu.sg, wumin@i2r.a-star.edu.sg).}
\thanks{Alvin Chan and Yew-Soon Ong are with the School of Computer Science and Engineering, Nanyang Technological University, Singapore and A*STAR, Singapore (Email: guoweialvin.chan@ntu.edu.sg, asysong@ntu.edu.sg).}
\thanks{Xiaoli Li is with Institute for Infocomm Research, A$^*$STAR, Singapore and the School of Computer Science and Engineering, Nanyang Technological University, Singapore (Email: xlli@i2r.a-star.edu.sg).}
}


%

\markboth{IEEE Computational Intelligence Magazine}%
{Shell \MakeLowercase{\textit{et al.}}: Bare Demo of IEEEtran.cls for IEEE Journals}
%

\maketitle

\begin{abstract}


Artificial Intelligence (AI) is a fast-growing research and development (R\&D) discipline which is attracting increasing attention because of its promises to bring vast benefits for consumers and businesses, with considerable benefits promised in productivity growth and innovation. To date it has reported significant accomplishments in many areas that have been deemed as challenging for machines, ranging from computer vision, natural language processing, audio analysis to smart sensing and many others. The technical trend in realizing the successes has been towards increasing complex and large size AI models so as to solve more complex problems at superior performance and robustness. This rapid progress, however, has taken place at the expense of substantial environmental costs and resources. Besides, debates on the societal impacts of AI, such as fairness, safety and privacy, have continued to grow in intensity. These issues have presented major concerns pertaining to the sustainable development of AI. In this work, we review major trends in machine learning approaches that can address the sustainability problem of AI. Specifically, we examine emerging AI methodologies and algorithms for addressing the sustainability issue of AI in two major aspects, \textit{i.e.}, environmental sustainability and social sustainability of AI. We will also highlight the major limitations of existing studies and propose potential research challenges and directions for the development of next generation of sustainable AI techniques. We believe that this technical review can help to promote a sustainable development of AI R\&D activities for the research community.

\end{abstract}


%
\IEEEpeerreviewmaketitle

\epstopdfsetup{outdir=./}

\section{Introduction}
%
%
%
%

\begin{figure}[!hpbt]
    \centering
    \includegraphics[width=0.5\textwidth]{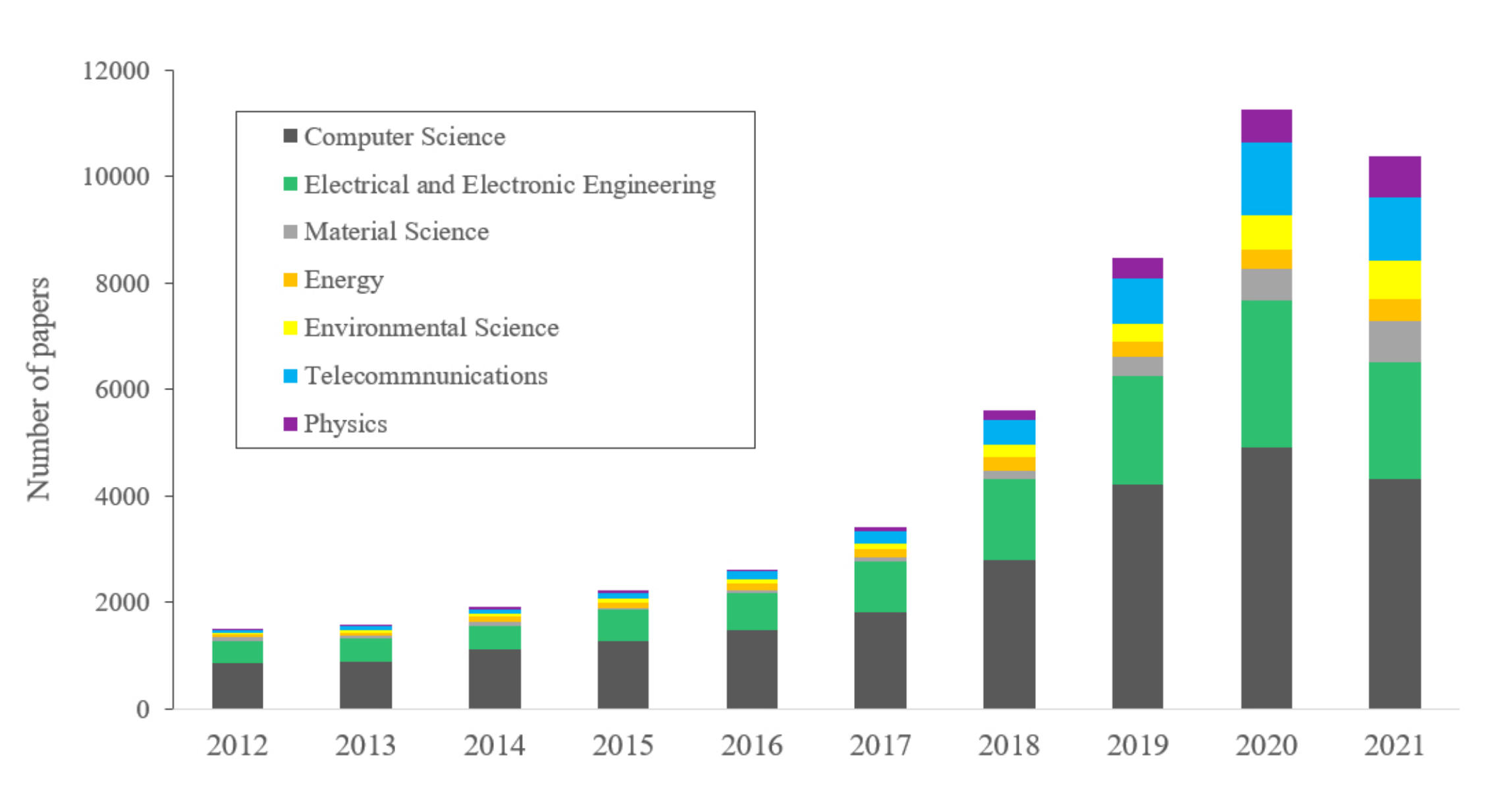}
    \caption{The trend of AI papers published in different disciplines for the past 10 years. Data are obtained from Web of Science as at Nov 22, 2021. }
    \label{fig:trend}
\end{figure}

Artificial Intelligence (AI) is intelligence demonstrated by computers or machines as opposed to the natural intelligence displayed by humans. It is capable of performing tasks that typically require human intelligence, even surpassing humans in many challenging tasks that have a large amount of data to learn from. In recent years, it has achieved remarkable successes in many areas, such as computer vision \cite{wang2020deep}, natural language processing \cite{devlin2018bert}, recommendation systems \cite{khanal2020systematic}, intelligent sensing \cite{chen2021generalization}, condition monitoring systems \cite{zhao2019deep}, advanced Web search engines \cite{chau2008machine}, understanding human speech \cite{amberkar2018speech}, automated decision-making \cite{leonetti2016synthesis}, competing at the highest level in strategic game systems \cite{silver2016mastering}, etc. Clearly, AI can play important roles in improving human work productivity and efficiency, reducing operational costs in many time-consuming and labour-intensive applications across different industry verticals and government agencies. Notable examples of AI applications include assisting doctors on their diagnostic tasks, self-driving cars, monitoring the operating conditions of expensive equipment, recommending suitable products at the right time and location, and facilitating human communication and interaction via machine translation tools. From Fig. \ref{fig:trend}, we can clearly observe an increasing trend of AI in not only the original discipline of computer science, but also other application domains, including material science, telecommunications, physics, environmental science, and others. It is clear
that AI is having significant impacts on and changing even traditional areas. All these will bring about a new era of interactions and augmentations between AI and human activities. 

\begin{figure*}[!hpbt]
    \centering
    \includegraphics[width=0.9\textwidth]{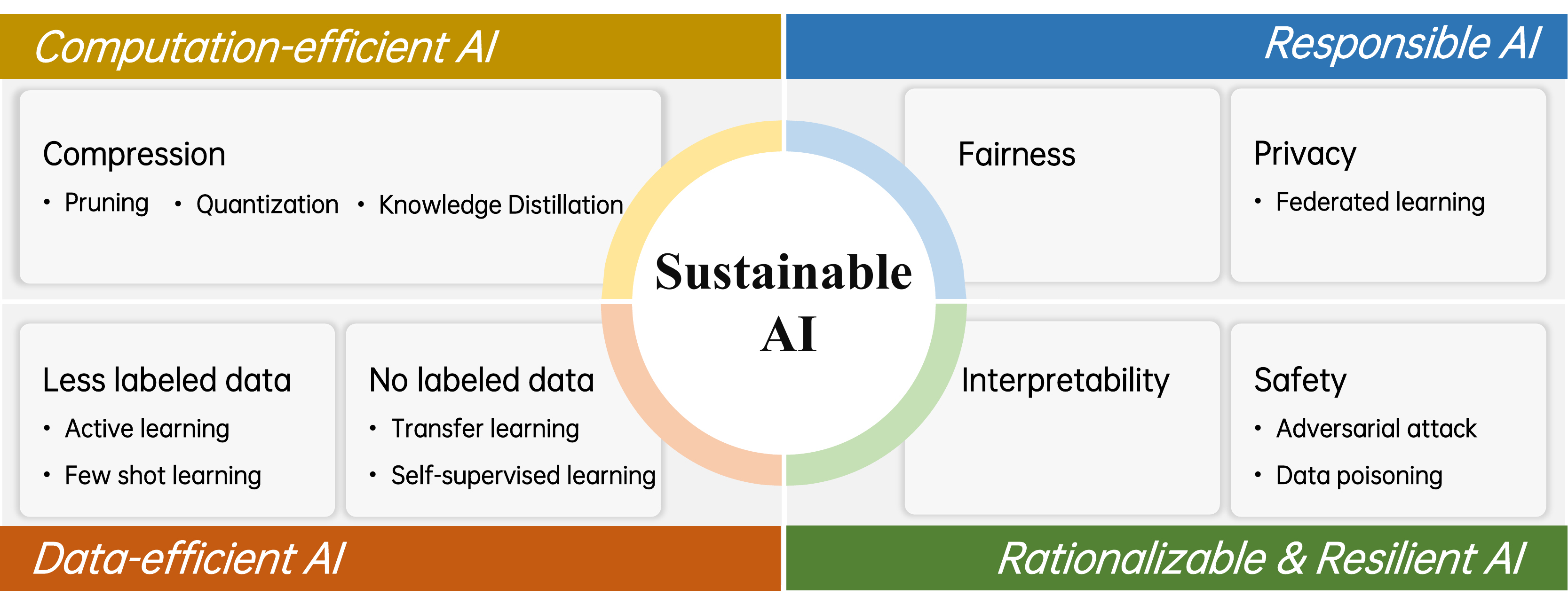}
    \caption{Overview of sustainable AI.}
    \label{fig:overview}
\end{figure*}

Beyond the technical achievements of AI, there have been increasing concerns about the sustainability of AI R\&D activities \cite{larsson2019sustainable,van2021sustainable}. For instance, researchers have reported that the carbon footprint of training a Transformer model is as high as 626,155 pounds of CO2 emission \cite{strubell2020energy}. Other concerns in AI ethics include fairness and privacy, which could significantly impede its progress and usage in life critical real-world applications if not appropriately taken into considerations \cite{hagendorff2020ethics}. In order to support continual progress and development of AI, the sustainability challenges must be taken into serious consideration and more efforts should be dedicated in this critical field. In this survey, our aim is to review the sustainability issue of AI from two major aspects, namely, to examine the recent and emerging technical research progresses made in the environmental sustainability and secondly focusing on the social sustainability of AI.

To date, several surveys that focused on sustainable AI \cite{larsson2019sustainable,kindylidi2021sustainability,van2021sustainable,vinuesa2020role} have been reported. Sustainable AI was defined in two perspectives \cite{van2021sustainable}, \textit{i.e.}, AI for sustainability (\textit{e.g.}, using AI to address climate change) and the sustainability of AI (\textit{e.g.}, reducing carbon emissions of AI models). 
In addition to the concepts above, the author in \cite{van2021sustainable} also discussed possible high-level strategies for sustainable AI. Larsson \textit{et al.} discussed sustainable AI in terms of the ethical, social and legal challenges for AI \cite{larsson2019sustainable}. They also summarized some recommendations for the development of sustainable AI, including AI governance, multi-disciplinary collaboration, transparency and accountability of AI. Kindylidi \textit{et al.} presented sustainable AI from the standpoint of customer protection, where the main focus has been on AI related policies \cite{kindylidi2021sustainability}. 
It is worth noting that existing works have focused on the strategic views for sustainable AI. \textit{None of the reviews to date, on the other hand, has considered the technical and algorithmic progresses towards achieving AI sustainability}. Considering that more and more attention has been devoted to these key technologies for achieving sustainable AI, this technical review attempts to fill this gap by reviewing emerging AI technologies (as shown in Fig. \ref{fig:overview}), highlighting how they contribute to the sustainable development of AI, discussing key challenges ahead and presenting potential future challenges and notable research directions worthy of investing time in.

The rest of the paper is organized as follows. We first introduce the detailed technical reviews from two perspectives of \textit{environmental sustainability} and \textit{social sustainability of AI} in Section II and III, respectively. Then, we provide detailed discussions and analyses in Section IV, and the future research challenges and directions are presented in Section V. Finally, we conclude this paper in Section VI.




\section{Environmental Sustainability of AI}

\begin{figure}[!hpbt]
    \centering
    \includegraphics[width=0.49\textwidth]{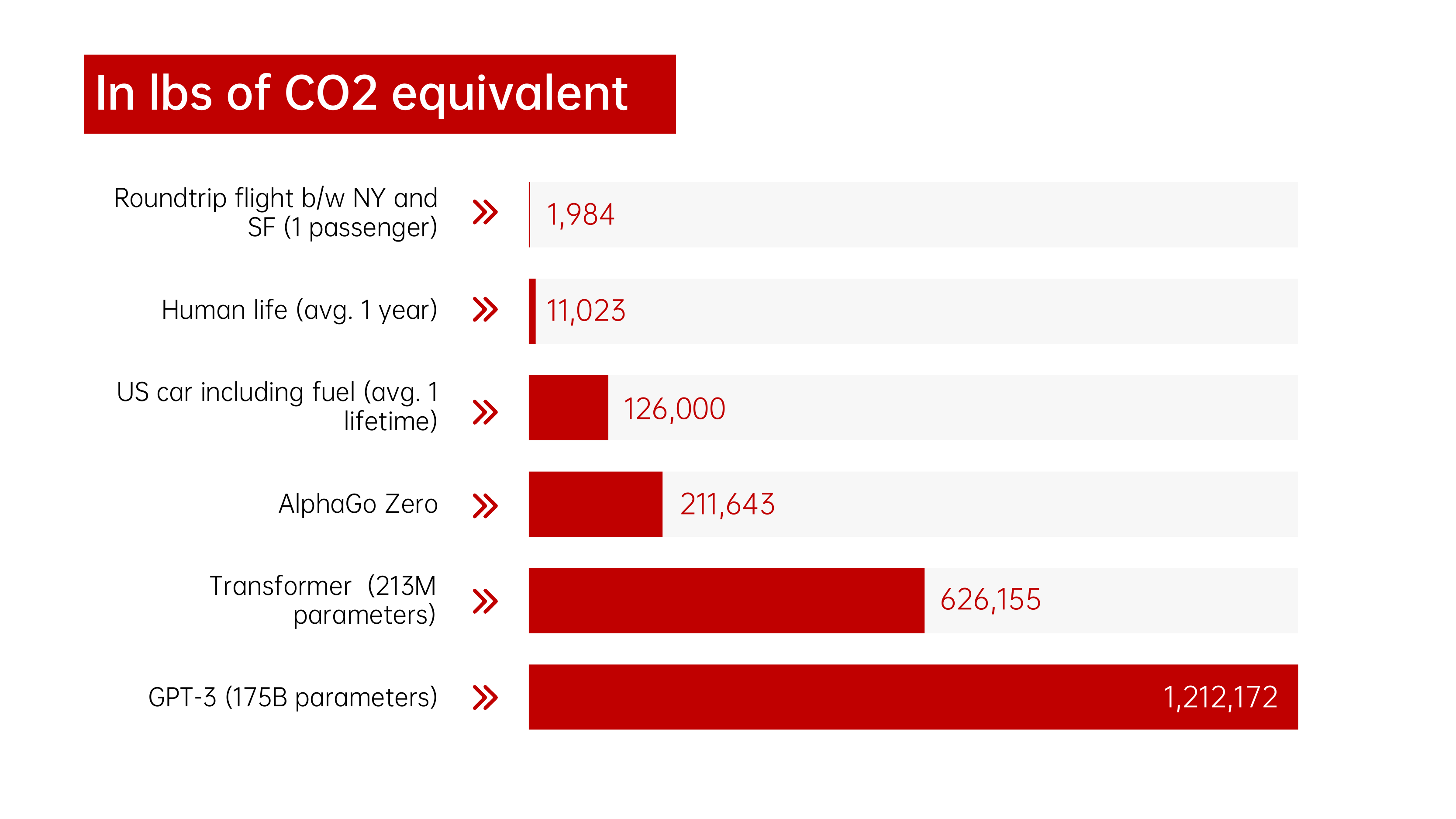}
    \caption{CO2 carbon footprint benchmarks. Source data from \cite{hao2019training,patterson2021carbon,van2021sustainable}.}
    \label{fig:co2}
\end{figure}

\textit{With increasing size and complexity of AI models, one big concern is its significant carbon footprint}. For example, a well-known study by \cite{van2021sustainable} reported that the process of training a GPT-3 model, \textit{i.e.}, a natural language processing model, can lead to alarming 1,212,172 pounds of CO2 emissions. This is roughly equivalent to the CO2 emissions of ten cars over their entire lifetimes. Fig. \ref{fig:co2} shows a clear trend of substantial CO2 carbon footprint of advanced AI models in comparison with daily CO2 emissions of human activities. This revelation clearly highlights the critical need to take environmental impacts into major consideration when developing AI.  

To address the sustainability impacts of AI, it is important to gain deep insights into how AI functions, and what technologies can be made available or developed to reduce its environmental impacts. To date, several major emerging techniques to address the environmental sustainability of AI have been investigated. Here, we divide these techniques into two core categories, namely, \textit{computation-efficient AI and data-efficient AI}.

\subsection{Computation-efficient AI}

\begin{figure}[!hpbt]
    \centering
    \includegraphics[width=0.5\textwidth]{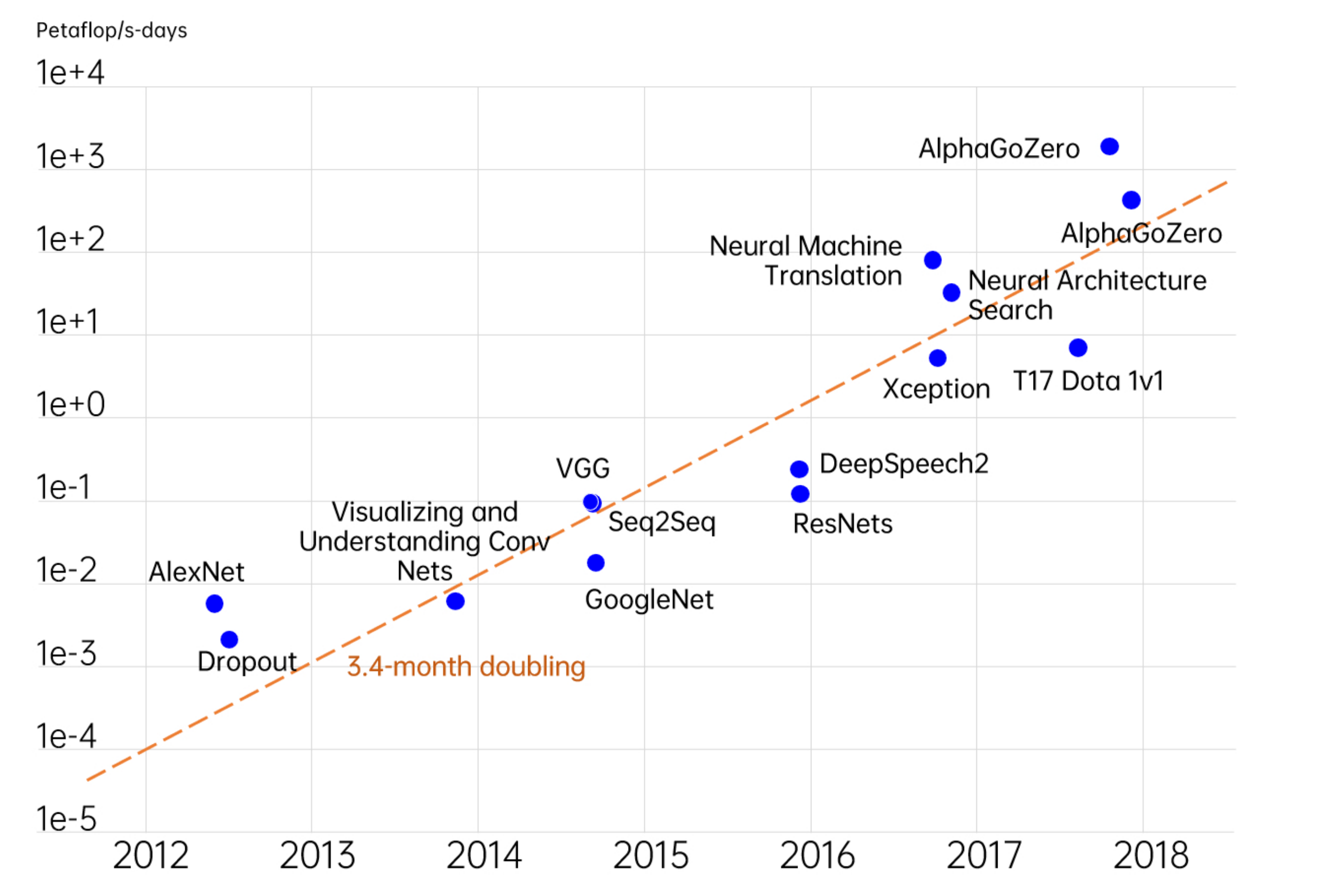}
    \caption{The computing requirements increased by 300,000 times from AlexNet \cite{krizhevsky2012imagenet} to AlphaGo Zero \cite{silver2017mastering}.}
    \label{fig:reAI}
\end{figure}

AI models have been shown to become more and more powerful along with the dramatic increase in their model sizes. According to Fig. \ref{fig:reAI}, the computation requirements of AI models will double every 3.4 months\footnote{https://openai.com/blog/ai-and-compute/}, which is much faster than Moore's law that claims to double every two years. The substantial size of AI models has a noteworthy impact on environmental sustainability, as more powerful platforms are required for their deployment and more energy will be consumed for real-time implementation. 

\textit{To reduce the environmental impact of AI, an effective way is to develop computation-efficient AI that requires less computational resources and at the same time generates smaller yet accurate models \cite{alemdar2017ternary}.} This has generally been established as compressed AI models. The compressed models could significantly reduce storage memory and energy for deployment, hence directly leads to lower carbon footprint. Many advanced techniques have been developed to build compressed AI models. We can divide them into three categories, \textit{i.e.}, pruning, quantization and knowledge distillation. We will discuss the details of each category in the following subsections. 




\subsubsection{Pruning}
\begin{figure}[!hpbt]
    \centering
    \includegraphics[width=0.5\textwidth]{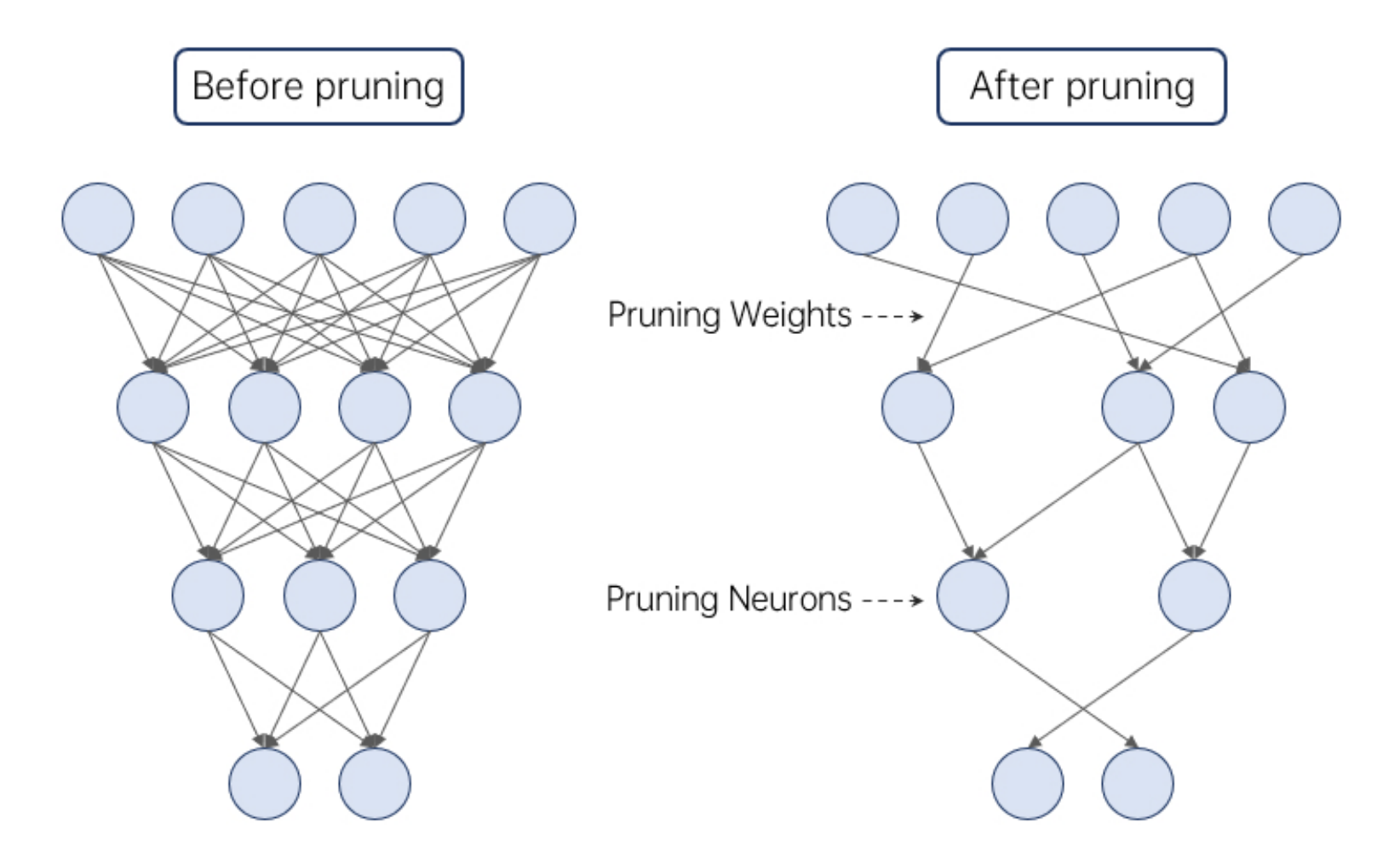}
    \caption{The process of network pruning.}
    \label{fig:pruning}
\end{figure}

With the consensus that AI models are typically over-parameterized \cite{lecun1990optimal, denil2013predicting,choudhary2020comprehensive}, network pruning serves to remove redundant weights or neurons that have least impact on accuracy, resulting in significant reduction of storage memory and computational cost. It is one of the most effective methods to compress AI models. Fig. \ref{fig:pruning} shows the process of pruning for a typical neural network based AI. 

Pruning criteria are essential for all pruning approaches. One of the simplest heuristics is to estimate importance in terms of the absolute value of a weight, called magnitude-based pruning. Zero-valued weights or weights within certain threshold are removed in \cite{han2015learning,li2016pruning}. Moreover, to zero the weights, regularization term is often adopted during training to increase weight sparsity \cite{hanson1988comparing}. However, these element-wise magnitude-based pruning methods may result in unstructured network organizations which are difficult to be compressed or accelerated without specialized hardware support \cite{liang2021pruning}. 

Structured pruning approaches, on the other hand, alleviate the above issue by pruning the network on the filter or layer level. For instance, Wen \textit{et al.} proposed a Structured Sparsity Learning (SSL) approach to regularize network structures (\textit{i.e.}, filters, channels and layers). Group Lasso is then used to zero out the grouped weights \cite{wen2016learning}. In addition to the weights, some other indicators have also been explored for pruning. Liu \textit{et al.} considered the scaling factor from batch normalization layer and pruned those channels of small scaling factor values \cite{liu2017learning}. Hu \textit{et al.} argued that certain activation function such as Rectified Linear Unit (ReLU) could generate numerous zeros and these zero activation neurons are redundant. Hence, they can be removed them without affecting the overall performance of the network \cite{hu2016network}. Moreover, some other researchers evaluated the importance (or saliency) of a neuron via its impact on the objective function. For instance, in Optimal Brain Damage \cite{lecun1990optimal} and Optimal Brain Surgeon \cite{hassibi1993second}, the second derivative (Hessian matrix) of the objective function with respect to the parameters is used to determine the redundant weights. Similarly, Lee \textit{et al.} introduced a saliency criterion based on the normalized magnitude of the derivatives in the objective function with respect to each weight \cite{lee2018snip}. 

With advanced pruning techniques, it is able to significantly reduce network size, which leads to reduced memory requirements and less computational cost in inference during model deployment stage \cite{JMAT2022}. However, some pruning techniques may result in more training efforts during model building stage, such as methods that are based on the procedures of train, prune and fine-tune/re-train \cite{huang2018learning,renda2020comparing}. A more efficient way is to perform pruning at initialization \cite{wang2020picking} or using sparse training, \textit{i.e.}, training under sparsity constraints \cite{mocanu2018scalable}.

\subsubsection{Quantization}

Another approach for computation-efficient AI is via quantization, where 32-bit floating-point parameters are quantized into lower numerical precision of 16-bit \cite{gupta2015deep}, 8-bit \cite{vanhoucke2011improving} or even 1-bit \cite{hubara2016binarized, rastegari2016xnor}. 
Using low-bit numerical representations not only reduce model storage cost but also result in significant speed-up during inference, as the most time-consuming process, floating point Multiply-Accumulate (MAC) operations, could be avoided. Jacob \textit{et al.} quantized both weights and activations into 8-bit integers for integer-arithmetic-only hardware \cite{jacob2018quantization}. Incremental Network Quantization (INQ) proposed in \cite{zhou2017incremental} converted full-precision weights into power-of-two values with lower precision. It allows the replacement of multiplication operations with bit-shift operations which are more efficient for hardware like Field-Programmable Gate Array (FPGA). 

Although quantization is generally applied to weights and activations, it can also be extended to gradient computations. DoReFa-Net \cite{zhou2016dorefa} quantized gradients to numbers with bit-width of less than 8 before back-propagation. It allows quantization of the network during training or fine-tuning. Instead of quantizing a single weight, vector quantization approaches focused on factorizing the weight matrix by weight clustering and sharing \cite{gong2014compressing, wu2016quantized }.

Instead of only using pruning or quantization for network compression, a combination of them have generally been reported to lead to higher compression rate. Han \textit{et al.} presented a three-stage pipeline which consists of pruning, weight sharing by vector quantization and Huffman encoding to compress a pre-trained model and it achieved state-of-the-art performance at that time \cite{han2015deep}. Both pruning and quantization are effective ways to reduce model size for efficient learning. One limitation is that they can only deal with models that share a common network architecture, which means compressing a cumbersome network to an efficient one with less weights (or neurons) and/or lower-precision. They cannot be used across heterogeneous networks, which can otherwise be addressed with the knowledge distillation technique. 



\subsubsection{Knowledge Distillation} 

Knowledge distillation (KD) is a flexible yet efficient approach to compress AI models. It aims to transfer the knowledge learnt from a cumbersome network to a compact one, which is also known as `Teacher-Student' learning framework \cite{wang2021knowledge}. With the knowledge from a teacher, a lightweight student network can achieve competitive performance as the cumbersome (teacher) network yet at higher computational-efficiency. The success of KD relies mainly on two key factors: the knowledge and distilling schemes. 

\begin{figure}[!hpbt]
    \centering
    \includegraphics[width=0.5\textwidth]{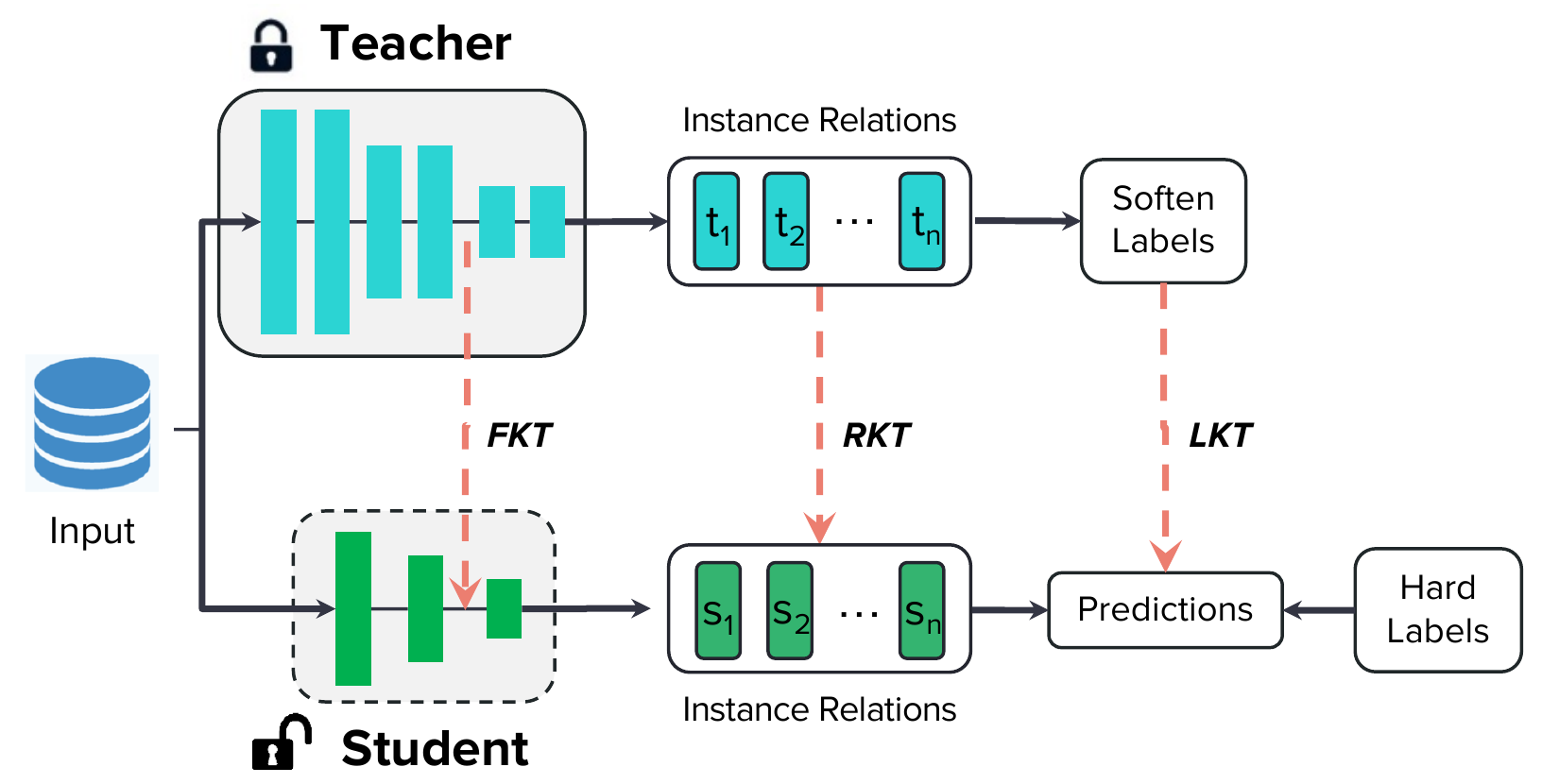}
    \caption{An illustration of knowledge distillation. FKT refers to Feature-based Knowledge Transfer, RKT refers to Relation-based Knowledge Transfer, and LKT refers to Logits-based Knowledge Transfer. }
    \label{fig:kd}
\end{figure}

The knowledge information can be categorized into three different types: logits-based, feature-based and relation-based knowledge \cite{gou2021knowledge}, which are depicted in Fig. \ref{fig:kd}. The vanilla KD approach leverages softened logits from a cumbersome teacher as the knowledge to guide the training process of a student \cite{hinton2015distilling}. Romero \textit{et al.} extended it by introducing the `hint' knowledge from intermediate layers of a teacher network, which can be termed as feature-based knowledge \cite{romero2014fitnets}. Instead of directly minimizing the discrepancy of feature maps between the teacher and the student, other advanced feature-based knowledge has also been explored. For instance, Zagoruyko and Komodakis regarded the teacher's attention maps derived from feature maps as the knowledge to be transferred \cite{komodakis2017paying}. 
Different from the aforementioned two knowledge types, relation-based KD methods emphasize the transfer of intra-relationships between data samples \cite{lee2019graph, park2019relational,tian2019contrastive} or correlations between feature maps from multiple intermediate layers \cite{lee2018self, yim2017gift}.

Along with these knowledge information, various distilling schemes have also been proposed. Although most of KD methods follow the conventional single-teacher single-student distilling scheme, some other promising schemes have also been explored in the literature. Firstly, the scheme of distilling informative knowledge from an ensemble of teachers instead of a single teacher could better enhance the student's generalization ability \cite{fukuda2017efficient,yuan2021reinforced}. 
Secondly, apart from distilling knowledge from large-scale networks, another research direction intends to boost the performance of a compact network by transferring the knowledge between networks which share identical architecture (also termed as self-knowledge distillation) \cite{furlanello2018born}. Furlanello \textit{et al.} proposed a sequential distilling scheme called Born-Again Networks (BANs) which take the model in earlier generation as the teacher and train a new initialized identical model \cite{furlanello2018born,zhang2018deep}. 
Thirdly, to better align feature representations between high-capacity and low-capacity networks, adversarial and contrastive distilling schemes are often adopted. To demonstrate the feasibility of transferring knowledge between two disparate network architectures, Xu \textit{et al.} exploited adversarial distilling scheme to transfer knowledge from a LSTM-based teacher to a CNN-based student \cite{xu2021kdnet}. 

Generally, KD methods are more flexible. The teacher and student networks in KD can be homogeneous or heterogeneous architectures, which is different from pruning and quantization. However, in most of cases, KD methods require more training efforts, as they need to train both the teacher and student networks. In fact, most of training efforts have been given to the teacher network which is often large in size and complex. To solve this issue, one way is to consider adopting a trained network as the teacher \cite{beyer2021knowledge}. Building a shared community for sharing trained AI models can significantly reduce repeated model training efforts, which can be an interesting direction for sustainable AI.



Computation-efficient AI can lead to compressed AI models which improve environmental sustainability of AI in two main aspects. First, compressed models can be deployed on resource-limited devices (\textit{e.g.}, embedded systems) with less memory and computational power. This also contributes to the feasibility of AI for various real-world applications. Second, compressed models are light-weighted and thus more efficient in real-time implementation, which can dramatically save energy and reduce carbon footprint. Although computation-efficient AI has clear merits towards sustainable AI, it may however have the drawback of increasing training effort. For example, the training of cumbersome teacher in knowledge distillation can be tedious. To solve this problem, a possible way is to build a shared community where researchers can upload and share their trained powerful but cumbersome models. Then, the tedious training process of teacher models can be avoided or at the very least not repeated. Nevertheless, compressed AI models are particularly useful in applications where they will be run for routine and long-term tasks, such as in condition monitoring where the system performs real-time equipment health condition monitoring for a few years \cite{xu2021kdnet}. In this case, the price that we pay to turn big models into smaller yet accurate models via tedious training process will be worthwhile in the long run. In summary, computation-efficient AI can contribute to environmental sustainability of AI with possible but addressable side effects by reducing the tedious training process, which requires more attention and continuous research efforts. 

\subsection{Data-efficient AI}

\textit{The other class of technology for environmental sustainability is to reduce the huge burden on data collection and annotation which require enormous human efforts and resources, as AI models generally require huge labeled datasets for model training.} Data-efficient AI addresses sustainable development of AI models by reducing human's manual efforts and resources. ImageNet \cite{deng2009imagenet}, for example, is said to have taken two and a half years of effort to be collected and annotated. However, most real-world applications would not have such luxuries, especially if a target use-case is niche and one is unable to use existing datasets. Compounding the problem is that the size of datasets directly affects training time and carbon footprint. In this subsection, we review advanced approaches that can help to reduce the data dependency, especially the size of labeled data. We divided these methodologies into two categories, \textit{i.e.}, using less labeled data and no labeled data.


\subsubsection{Less Labeled Data}

AI is a typical data-driven method and its performance correlates highly with sufficient amount of labeled data for model training. This huge burden on data collection and annotation will significantly impact on the environmental sustainability of AI. To use less labeled data, some progressive AI technologies have been developed, including active learning and few-shot learning. 

\paragraph{Active Learning}

\begin{figure}[!hpbt]
    \centering
    \includegraphics[width=0.5\textwidth]{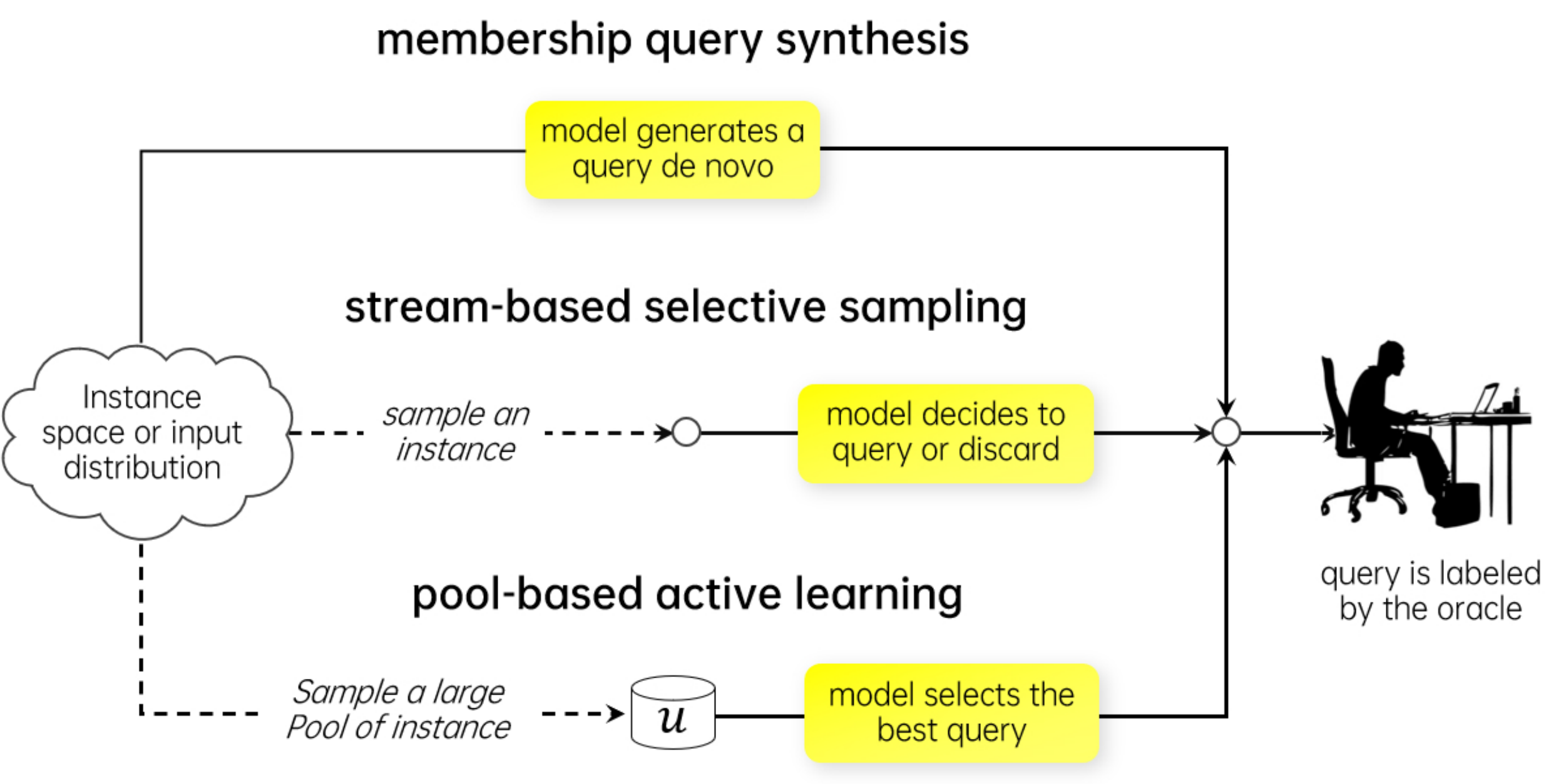}
    \caption{The procedure for active learning \cite{settles2009active}. }
    \label{fig:active}
\end{figure}

To use less labeled data for model training, active learning intends to find representative samples (\textit{e.g.}, a small portion of the entire dataset) to be annotated by the oracle (\textit{e.g.}, human annotator), such that a good supervised learning model can be trained with intelligently selected annotated samples \cite{ren2020survey}. Generally, active learning techniques can be divided into three different groups of membership query synthesis, stream-based selective sampling and pool-based sampling, 
as shown in Fig. \ref{fig:active}.

For the membership query synthesis method, learners can select any unlabeled samples from instance space or generated samples by the learners for annotation \cite{schumann2019active}. It is particularly effective for small datasets. However, the generated samples may be difficult for the oracle to annotate, which limits its feasibility in many real-world applications. Stream-based selective sampling sequentially sends a data sample to an active learner which will decide whether to annotate or discard the sample \cite{smailovic2014stream}. On the other hand, pool-based sampling approach aims to select samples for annotation from a pool of unlabeled samples. Hence, it is able to select more than one sample at a time, which can be more efficient.

Currently, pool-based sampling is the most widely used technique in active learning. Its key component is the sample selection strategy. The basic idea of selection strategies is to select samples that contain rich information and are highly representative and diverse \cite{ren2020survey}. A popular selection strategy is based on uncertainty sampling \cite{he2019towards}, which measures the uncertainty of samples based on techniques like least confidence and entropy \cite{siddiqui2020viewal}. Density based methods intend to query consecutive points from the instance pool to construct a core set for representing the distribution of feature space of original samples \cite{geifman2017deep}. Active learning can significantly reduce the efforts for data annotation as it only annotates representative and hard/ambiguous/uncertain examples. Besides the training can be more efficient as the selected samples represent a small portion of the entire dataset.

\paragraph{Few-shot Learning}

Few-shot learning is another class of approaches that aims to learn from less labeled data for training AI models \cite{wang2019few}. To solve few-shot learning problems, there are two main groups of methodologies. The first one is based on meta-learning, also known as learning to learn \cite{vanschoren2018meta}. Meta-learning techniques generally consist of a teacher model and a student model, where the teacher learns how to optimize the student while the student learns how to perform downstream tasks. This can be seen in \cite{finn2017model} where the outputs of the teacher models were used to train the student model by optimizing and selecting the parameters from a higher dimensional space due to the limited training samples. Generally, initial meta-learners are biased towards existing tasks in parameter optimization, but Jamal and Qi \cite{jamal2019task} demonstrated the possibility of using an agnostic meta-learner to overcome those limitations. Another promising work is presented by Hou \textit{et al.} \cite{hou2021meta} where attention modules for meta-learners were used to achieve the task-specific initialization of base models for fast adaptation in meta-learning.

Another popular approach for few-shot learning is metric learning. The general direction is to compare the embeddings of the support (few-shot training samples) and query (test samples) sets. Examples of such work include the Siamese \cite{koch2015siamese} and Triplet \cite{hoffer2015deep} networks which compared two embeddings at one time, and the Matching \cite{vinyals2016matching}, Prototypical \cite{snell2017prototypical} and Relation \cite{sung2018learning} networks which compared across multiple embeddings from support and query sets. A distinguishing feature across metric learning approaches is the distance measures for embeddings. The Triplet, Matching and Prototypical networks adopt pre-defined distance measures whereas the Siamese and Relation networks learn the distance measures by using neural networks.

Similar to active learning, few-shot learning can significantly reduce the amount of labeled data for model training, which saves a lot of resources required for data collection and annotation, thus reducing carbon footprint in these tedious processes. However, few-shot learning may not be able to reduce training effort if learning with complex algorithms, like meta-learning which requires additional training efforts, or metric learning where comparing high-dimensional embeddings may be time-consuming. These negative impacts must be considered and evaluated under the perspective of environmental sustainability. 

\subsubsection{No Labeled Data}

More advanced AI techniques intend to release the burden of data annotation by learning AI models via data from related tasks or unlabeled data. We will introduce two typical techniques, \textit{i.e.}, transfer learning and self-supervised learning, and analyze their contributions to environmental sustainability. 

\paragraph{Transfer Learning}

\begin{figure}[!hpbt]
    \centering
    \includegraphics[width=0.48\textwidth]{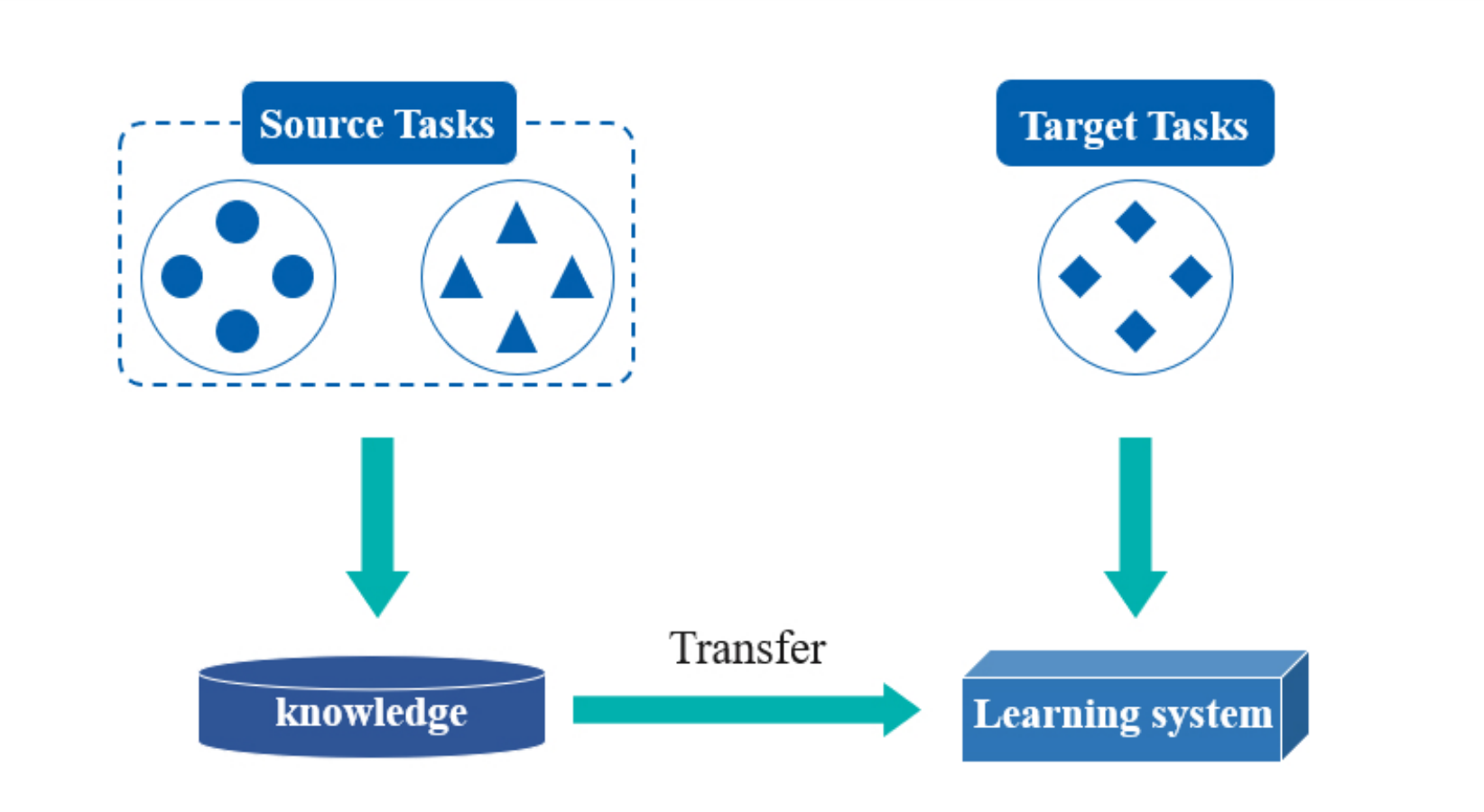}
    \caption{Learning process of transfer learning \cite{pan2009survey}. }
    \label{fig:TL}
\end{figure}

Transfer learning aims to solve a given task (denoted as target domain) with labeled data from related ones (denoted as source domain) \cite{pan2009survey}, as shown in Fig. \ref{fig:TL}. Note that generally the target domain only contains unlabeled data. The main idea in transfer learning is to minimize the domain difference between the source and target domains. There are two types of approaches to minimize domain difference. The first one is \textit{distance-based methods}, where a distance function, such as MMD \cite{ddc,HoMM2020}, CORAL \cite{sun2017correlation}, etc., is defined to measure the domain difference, and it will be minimized during training, such that the gap between the source and target domains is minimal \cite{rahman2020minimum}. Then, the source classifier/regressor trained using the labeled source domain data can be used for the classification/regression task in the target domain. The second type is \textit{adversarial-based methods} which are inspired by generative adversarial network (GAN) \cite{adda,DANN,wudeep}. They design a discriminator to distinguish the features generated by the feature encoders of source and target domains. At the same time, the feature encoders of the source and target domains will be trained to push the features from both domains to be similar such that the discriminator cannot separate them \cite{wang2020self}. 

Transfer learning which assumes the availability of a labeled source domain (related to target) is a common setting in real applications. For example, when performing an image classification task with no available labeled data, an effective way is to use transfer learning with public datasets, like ImageNet \cite{deng2009imagenet}. However, it still requires the collect of unlabeled data in the target domain for the adaptation. Besides, the training effort will also increase as the model will be trained with both source and target domain data. In this case, there will be a trade-off between the efforts of data annotation and model training. If we are able to quantify annotation efforts in terms of carbon footprint, it may help us to better evaluate on whether transfer learning can actually contribute to environmental sustainability. 

Note that there is a special case of transfer learning, named fine-tuning \cite{guo2019spottune}, which assumes the availability of few labeled data in the target domain. Then, fine-tuning aims to train just few layers of the learnt network from the source domain (also known as pre-trained model) by using few labeled samples in the target domain. Suppose that the pre-trained model is only obtained from a shared community without any training efforts, fine-tuning can significantly contribute to environmental sustainability as it does not require large datasets (either labeled or unlabeled) and can dramatically reduce training costs (only parameters from few layers will be trained). Here, we emphasize again the importance of creating a shared community for trained AI models towards the development of sustainable AI.


\paragraph{Self-supervised Learning}

Another popular technique that does not use labeled data is self-supervised learning, which is able to learn representations directly from unlabeled samples. It consists of two main approaches to achieve this objective. The first one is contrastive learning which learns features by identifying similar data samples and dissimilar ones. SimCLR \cite{chen2020simple} presented a typical contrastive learning framework. It learns representations by maximizing the similarity of two different augmented views of one sample and minimizing the similarity of augmented views of different samples. It has shown that the number of negative pairs will significantly affect the performance of representation learning. However, a large number of negative pairs require huge memory. To address this issue, MoCo \cite{he2020momentum} designed a memory bank to store and reuse representations of data. MoCo-V2 \cite{chen2020improved} is an updated version of MoCo, which also borrowed the ideas of project head and augmentation from SimCLR. BYOL \cite{grill2020bootstrap} claimed a new state-of-the-art performance for representation learning without using any negative samples. It is composed of two networks, \textit{i.e.}, online and target networks, that interact and learn from each other. SimSiam \cite{chen2021exploring} simplified the architecture of BYOL and developed a novel method of stop gradient to prevent it from collapsing during learning, which has been shown to be effective in representation learning. 

Another widely used approach for self-supervised learning is based on pretext tasks. The objective is to design pretext tasks as supervised signals for learning representative features. For instance, a popular way is to predict the relative positions of image patches \cite{doersch2015unsupervised}, which will promote the model to recognize objects and patches, such that good feature representations can be learnt. Alternatively, it can predict the rotation degrees of images, which define four degrees for prediction, \textit{i.e.}, 0, 90, 180, and 270 \cite{komodakis2018unsupervised}. Noroozi and Favaro proposed a pretext task by solving jigsaw puzzles of images \cite{noroozi2016unsupervised}. In this challenging task, the model tends to learn a feature mapping of objects in images and their correct spatial relations. Other visual pretext tasks include colorization \cite{larsson2016learning}, image inpainting \cite{yu2018generative}, perturbations \cite{carr2020shuffle}, etc. In addition to visual applications, many pretext tasks have been proposed in natural language processing (NLP), such as predicting center words or neighbor words in a sentence \cite{mikolov2013efficient}. The powerful BERT \cite{devlin2018bert} has designed several pretext tasks. For example, it masks a word in a sentence and predict it. Other pretext tasks include next sentence prediction, sentence permutation, document rotation, etc. 

Self-supervised learning can extract useful information from the data by contrasting between samples with their augmentations or defining pretext tasks as supervised signals. It totally releases the effort of data annotation. However, the creation of augmented samples or pretext tasks can lead to additional training burden, which will consume more computational power. Due to the lack of ground truth labels, self-supervised learning generally requires more training iterations to converge. Overall, the current self-supervised learning avoids the cost and efforts in data annotation but dramatically increases the training burden. Although, self-supervised learning is a very promising research area in AI, it requires further research to reduce its training effort for sustainable development of AI. 


\section{Social Sustainability of AI}


In addition to the environmental impact of AI, another key issue that requires a lot of attention is its social impact. Here, social sustainability of AI means issues pertaining to users and interactions between users and AI models. Specifically, we consider two technical aspects of social sustainability of AI, namely, Responsible AI and Rationalizable \& Resilient AI. In responsible AI, the key issues of AI fairness across different users and user data privacy will be explored. 
While in rationalizable \& resilient AI, we discuss interpretability and safety of AI models, which are crucial in cultivating greater acceptance of AI models \cite{ong2019air}. 


\subsection{Responsible AI}

It can be foreseen that AI systems will soon play key roles in various aspects of human life. As such, AI ethics must be considered when designing AI systems. 
In this technical review, we discuss two major aspects of responsible AI, \textit{i.e.}, fairness and data privacy. The fairness of AI is to eliminate the biases in AI models towards different users (\textit{e.g.}, with different gender, race, religion, etc.), where many research efforts have been directed to this critical area. Another key aspect is about the data privacy of users. Without careful and secure protection of data privacy, users will not be willing to share their data for the training of AI models. Here, we will review emerging AI techniques for addressing AI fairness and privacy issues in the following paragraphs. 

\subsubsection{Fairness}

As various AI models are continually making great achievements in sociotechnical systems, they may suffer from unexpected social implications, including social bias towards race, gender, poverty, disabilities, etc. A classic example is a software system named COMPAS used by American courts to judge the risk of an offender recommitting another crime \cite{green2019disparate}. In this system, the black people are always assigned with higher risk than other groups even though other inputs are similar. Such biases or discriminations can also be found in facial recognition systems \cite{raji2019actionable} and recommendation systems \cite{schnabel2016recommendations}. Therefore, it would be crucial to promote AI fairness to mitigate these bias and discrimination issues for social harmony and justice. 

Unfairness generated by AI models usually arises from two types of biases, namely, biases in data and the algorithmic biases \cite{mehrabi2021survey}. The most common bias in data is that the data contains sensitive features (also called protected attributes), \textit{e.g.}, race and gender. The use of such features in AI models will lead to direct or indirect discrimination in some specific tasks. There are some other biases in data, such as conformity bias and popularity bias, in recommendation systems \cite{chen2020bias}. Conformity bias refers to that the users tend to behave similarly to their friends, and such behaviors do not always reflect their true preferences. Popularity bias means that the popular items tend to be exposed more. Popular items would thus be recommended more frequently and less popular items tend to get limited attention. Different from data bias, algorithmic bias is induced purely by AI algorithms. Do refer to \cite{mehrabi2021survey, chen2020bias} for more details about different types of biases. 

To mitigate the above biases, various metrics to quantify the fairness have been developed in recent years \cite{mehrabi2021survey, caton2020fairness}. Some of most widely used definitions of fairness include \textit{demographic parity} \cite{zemel2013learning}, \textit{equal opportunity} \cite{hardt2016equality}, and \textit{fairness through awareness/unawareness} \cite{dwork2012fairness}. For example, demographic parity requires the predictions to be independent of the sensitive feature. For different types of fairness, please refer to \cite{mehrabi2021survey, caton2020fairness, dwork2012fairness} for more details. In addition, fairness definitions can fall under two categories, namely, individual fairness and group fairness. Individual fairness requires that the model generates similar predictions for similar individuals, while group fairness requires that the model treats different groups equally. Therefore, demographic parity and equal opportunity are group fairness, while fairness through awareness/unawareness is individual fairness.



\begin{figure}[!hpbt]
    \centering
    \includegraphics[width=0.5\textwidth]{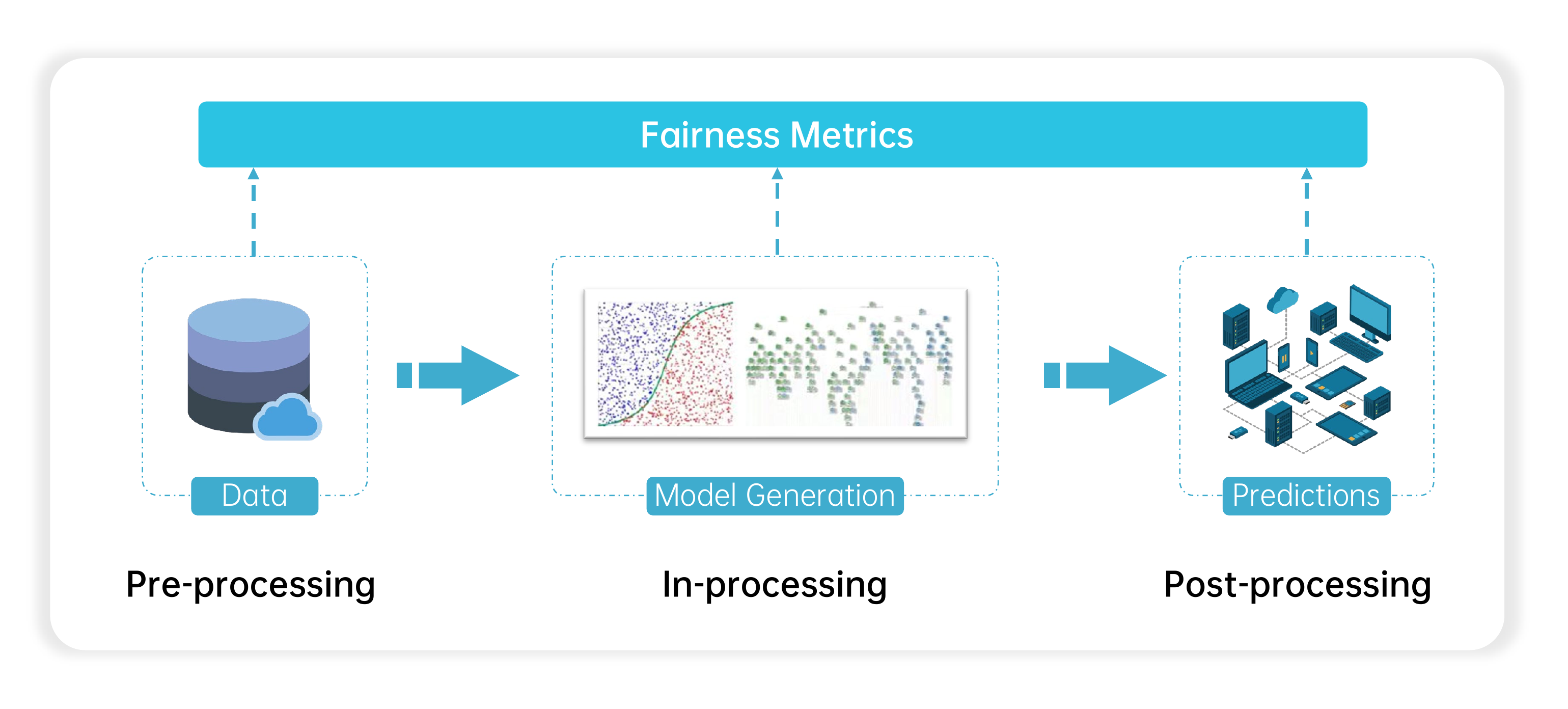}
    \caption{Three main categories of approaches to improve fairness.}
    \label{fig:fair}
\end{figure}

To achieve AI fairness, various approaches have been proposed \cite{bellamy2018ai, mehrabi2021survey, caton2020fairness}. These approaches can be categorized as pre-processing, in-processing and post-processing methods as shown in Fig. \ref{fig:fair}. Pre-processing approaches focus on transforming the data to remove biases and discriminations. Common pre-processing approaches include variable blinding \cite{kamiran2012data, feldman2015certifying}, relabelling \cite{jiang2020identifying} and data reweighing \cite{kamiran2012data}. Moreover, adversarial learning is also applied to generate high-quality fair training data \cite{xu2019achieving}. In-processing methods include fairness constraints during model training and maximize both performance (\textit{e.g.}, accuracy) and fairness. For example, fairness constraints are considered when optimizing the accuracy for classification \cite{zafar2017fairness} and regression tasks \cite{komiyama2018nonconvex}. Post-processing methods aim to improve the model fairness by applying transformations (\textit{e.g.}, thresholding \cite{valera2018enhancing}) to the model outputs. 

AI fairness, covering both the fairness metrics and fair AI models, is a very active research topic. However, there are still many open issues and future opportunities in this area. First, as fairness is a multi-faceted concept, various fairness metrics have been proposed \cite{narayanan2018translation}, and some of them may be incompatible with the others. To address this incompatibility issue, it would be necessary to choose suitable fairness metrics based on the context and characteristics of current AI applications. 
Second, when improving the fairness, other aspects of AI models may be affected, \textit{i.e.}, fairness-utility trade-off. It is thus important to achieve a balance between fairness and utility factors (\textit{e.g.}, accuracy \cite{wang2021understanding}, and privacy \cite{khalili2021improving}).

\subsubsection{Privacy}

Another key aspect in responsible AI is data privacy. Generally, the training of AI models requires a large dataset which may be held by different parties/organizations. Data sharing across parties/organizations will lead to privacy concerns which would limit the cooperation for sustainable development of AI. To address this issue, federated learning (FL), which trains an algorithm across multiple decentralized edge devices or servers holding local data samples without exchanging their data samples, can effectively address these security and privacy issues \cite{yang2019federated,chen2017workload}. The authors in \cite{yang2019federated} presented a comprehensive introduction of FL, including 1) sample-based and 2) feature-based federated learning algorithms. They also discussed the differences of FL with distributed learning and edge computing, as well as recent works/applications of FL. Next, we briefly review sample-based and feature-based federated learning algorithms.


\begin{figure}[!hpbt]
    \centering
    \includegraphics[width=0.48\textwidth]{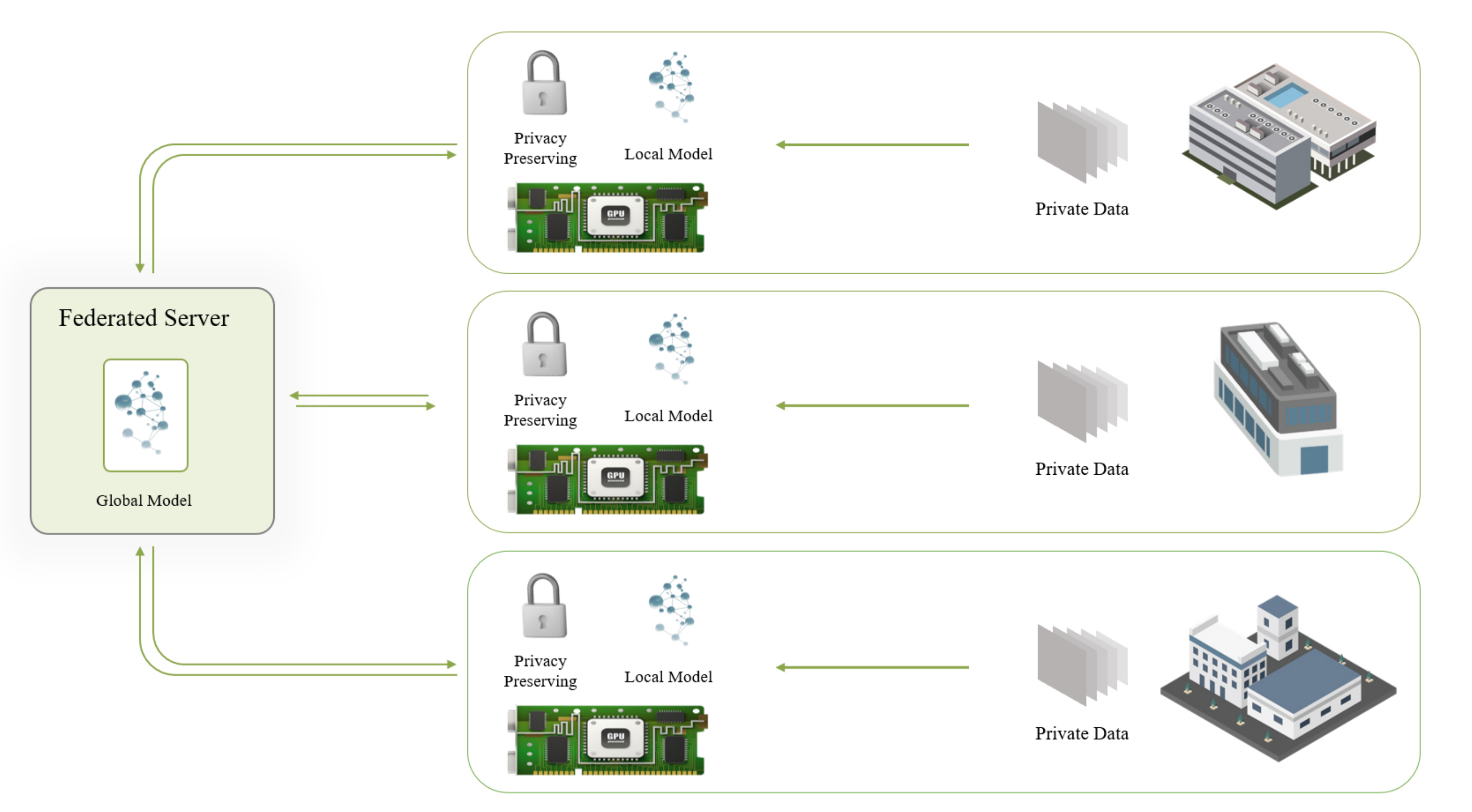}
    \caption{An illustration of sample-based federated learning. Each party learns a local model based on its own data and then sends the local model to a cloud server. The server aggregates the model parameters from all the parties and then shares the aggregated global model to each party for updating their respective local models. }
    \label{fig:FL}
\end{figure}

\textbf{Sample-based federated learning}, also called horizontal federated learning, is proposed for the scenarios where different parties have different data samples with the same features. Fig. \ref{fig:FL} shows a typical framework for sample-based FL, where three hospitals have different patients with the same health and medical features. A federated learning framework called MOCHA \cite{smith2017federated} was proposed to allow multiple sites/parties to complete their individual tasks while sharing the knowledge and preserving data privacy. Meanwhile, MOCHA addressed the issues of high communication cost and fault tolerance in federated learning. The authors in \cite{konevcny2016federated} have proposed federated optimization methods, \textit{e.g.}, federated stochastic variance reduced gradients (federated SVRG). These optimization methods can improve communication efficiency and facilitate the training of high-quality centralized models based on the data from multiple mobile clients. The authors presented a framework for federated learning of deep neural networks based on iterative model averaging called FedAvg \cite{mcmahan2017communication}. The researchers in \cite{li2019convergence} further derived a bound for the convergence rate of FedAvg, showing a trade-off between communication efficiency and convergence rate. Konecny \textit{et al.} \cite{konevcny2016federated} proposed two ways to reduce uplink transmission costs, \textit{i.e.}, structured and sketched updates. And the main difference of the two methods is on learning updates from variables or models. Tran \textit{et al.} \cite{tran2019federated} investigated two trade-offs of using FL over wireless communications, \textit{i.e.}, computation versus communication latencies and FL time versus user equipment energy consumption. The authors derived the optimal solution by characterizing the closed-form solutions. Wang \textit{et al.} \cite{luping2019cmfl} proposed a CMFL framework to identify irrelevant client updates. The global tendency was used as the overall information for all the clients to check the alignment for updates.

\textbf{Feature-based federated learning}, also called vertical federated learning, is proposed for the scenarios where the data are vertically split (\textit{i.e.}, by features) over multiple parties. Assume that there are two parties (\textit{i.e.}, data providers A and B) and a third-party collaborator (\textit{i.e.}, the cloud server C). The training process of feature-based federated learning can be described with the following four steps \cite{yang2019federated}. First, cloud server C creates encryption pairs and sends a public key to A and B. Second, A and B encrypt and exchange the intermediate results for gradient and loss calculations. Third, A and B compute encrypted gradients and send the encrypted values to C. Fourth, C decrypts and sends the decrypted gradients and loss back to A and B, so that A and B can update the model parameters respectively. The authors presented a logistic regression classifier for feature-based federated learning where the data are vertically split \cite{hardy2017private}. Their solution includes privacy-preserving entity resolution and federated logistic regression over messages encrypted with an additively homomorphic scheme. The authors in \cite{yang2019parallel} further proposed a solution for parallel distributed logistic regression in feature-based federated learning without using a third-party coordinator. Hence, the proposed solution can reduce the complexity of the system and allows any two parties to train a joint model without the help of a trusted coordinator. A lossless privacy-preserving tree-boosting system named SecureBoost \cite{cheng2021secureboost} was proposed for feature-based federated learning. SecureBoost first conducts entity alignment under a privacy-preserving protocol and then constructs boosting trees across multiple parties through vertical federation. A privacy-preserving feature-based federated learning for tree-based models named Pivot was proposed in \cite{wu20privacy}. Similar to the solution in \cite{yang2019parallel}, Pivot also does not rely on any trusted third-party coordinator and thus provides better privacy protection and lower system complexity. 


More recently, several advanced federated learning algorithms have been proposed and applied for more challenging scenarios (\textit{e.g.}, cross-domain scenarios to address both the domain shift issue and the data privacy issue.) \cite{yang2019federated, li2021survey}. Liu \textit{et al.} \cite{liu2020secure} proposed a secure federated transfer learning (FTL) system which enables complementary knowledge to be transferred across domains without compromising the data privacy. In particular, both homomorphic encryption and secret sharing were used in federated learning for privacy preservation. Federated domain adaptation \cite{peng2019federated} and federated domain generalization \cite{liu2021feddg} were proposed to address the domain shift issue and generalize the federated model to the target domains (\textit{e.g.}, new parties or sites). However, there are still many challenging issues and open questions in federated learning \cite{kairouz2019advances} to be addressed in the future. For example, federated learning has primarily considered supervised learning tasks where labels are available for each data provider. Extending federated learning to other machine learning paradigms, including reinforcement learning, semi-supervised and unsupervised learning, is interesting and challenging. In addition, it is also very promising to consider other ethical values (e.g., fairness, safety, interpretability, etc.) in federated learning systems.


\subsection{Rationalizable \& Resilient AI}

To promote the acceptance of modern AI systems for social sustainability, the mechanism of AI models is required to be more rationalizable, in other words, possess the ability to be rationalized (interpretable) \cite{ong2019air}. This is also referred to as the interpretability of AI models. Another aspect to encouraging acceptance is the resilience of AI models, \textit{i.e.}, their ability to preserve adequate performance even under extreme cases, such as adversarial attacks (input perturbations) \cite{ong2019air}, which is also known as the safety of AI. In the following, we will review technical details for achieving AI interpretability and safety. 


\subsubsection{Interpretability}

It is well known that most of the AI systems are black boxes, where it is difficult to explain how they work, especially to the general public and users. Users may not be able to trust AI models if they cannot understand their decision-making process. Especially for critical domains, like finance, medication, healthcare and self-driving cars, it is unacceptable without the interpretability of AI models. Recently, more and more attention has been focused on how to explain the decisions of AI, also known as explainable AI. Many innovative explainable AI techniques have been developed in the past few years. We can roughly divide them into three categories, namely 1) model-based, 2) model-agnostic, and 3) example-based methods. 

\textbf{Model-based methods} mainly explore AI models that can be explained. Typical explainable AI models includes linear models, \textit{e.g.}, linear regression and logistic regression, and tree-based methods, \textit{e.g.}, decision tree and decision rule \cite{molnar2020interpretable,xlli2013}. For instance, the weight of a feature in linear regression represents the importance of the feature, which can be useful in explanation. The if-then rules in a decision rule algorithm clearly indicate the relationship between inputs and outputs of the model. Domain experts, even without deep AI knowledge, can help to check whether the rules that are automatically learnt from data do really make sense or can be further refined. 

\textbf{Model-agnostic solutions} are more general in explainable AI, which can be used for any AI models. A typical method is based on partial dependence plot (PDP) which shows the marginal effect of features on prediction results of an AI model \cite{friedman2001greedy}. LIME \cite{ribeiro2016should} intended to perturb inputs and see the changes of model outputs, which can be used to explain the relationship between inputs and outputs. Another popular method named Anchors \cite{ribeiro2018anchors} tried to emphasize a part of inputs that are sufficient for an AI model to give predictions. 

\textbf{Example-based methods} use a set of examples from a dataset to explain the underlying behaviors of AI models. Counterfactual explanations describe how an instance changes to significantly influence its prediction \cite{verma2020counterfactual}. By designing counterfactual samples, we are able to know how the model produces certain predictions and how to explain individual predictions. Another example-based method leverages prototypes and criticisms in which prototypes select representative samples from data and criticisms select samples that are not well represented by prototypes \cite{kim2016examples}. The way of selecting influential samples is to identify training instances that are most critical to the model or predictions \cite{koh2017understanding}. Through analyzing these influential samples, one can better understand the model's behavior, which leads to better explainability.

Although various techniques have been developed for explaining AI models, existing solutions can only deal with simple models or simply explain the relationships between inputs and outputs. The underline behaviour of complex deep AI models, \textit{e.g.}, what has actually been learnt in each layer of a deep neural network, is still a mystery and more research efforts should be given to this crucial area.  

\subsubsection{Safety}

Deep learning-based AI systems have witnessed growing adoption due to their impressive performances and even have recently been expanded into critical applications where stakes are high, such as self-driving vehicles where the safety of such systems is paramount. Notably, some motor accidents have already been connected to self-driving systems, highlighting the immense importance of AI safety. Apart from implications on human life, safe AI builds trust in society that is pivotal to its sustainable adoption by the people. Hence, it is key to consider possible threats where the system fails to perform according to its users’ expectations. Recent research has shown that an adversary can undermine the safety of the system by targeting either the AI’s inference or training phase, by means of adversarial examples and data poisoning respectively.


\begin{figure}[!hpbt]
    \centering
    \includegraphics[width=0.5\textwidth]{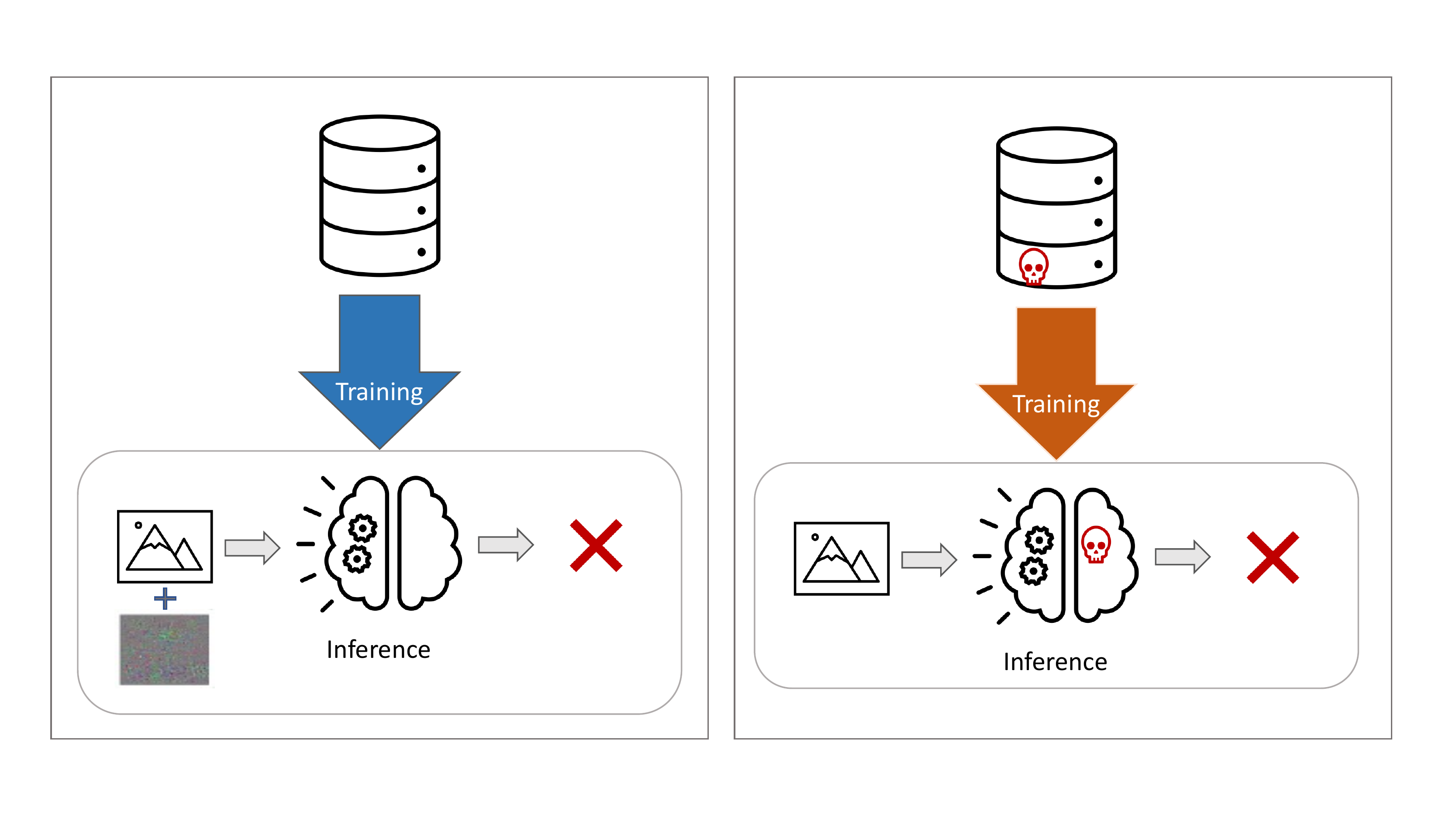}
    \caption{Illustrations of how (left) adversarial examples and (right) data poisoning can fool an image classifier by targeting the training and inference phases, respectively.}
    \label{fig:safe_ai}
\end{figure}

\textbf{Adversarial Examples:}
Much like visual blind spots in human vision, visual deep learning systems are vulnerable to imperceptible perturbations called adversarial examples \cite{szegedy2013intriguing}. In a self-driving car example, small perturbations can be added to an image of a road sign to steer the car's image classifier towards classifying the sign as a different object (\textit{e.g.}, a high-speed sign when it is a stop sign) with high confidence, which could result in a catastrophic outcome. Such perturbations can be crafted as the input image is fed into the AI model, through the access of the model's prediction probability and gradient information \cite{papernot2018cleverhans,croce2020reliable}. This information could be exploited by an adversary to compute the subtle corruption to the input image that can steer the model's prediction towards a target class. Apart from the image domain, the threat of adversarial examples exists in the audio \cite{carlini2018audio,wenger2021hello,zelasko2021adversarial} and natural language domains \cite{jia2017adversarial,xu2020adversarial,tan2020s}.

Fortunately, defenses against adversarial examples have emerged. Among them, the most effective defenses are based on a simple concept of \textit{adversarial training} (AT) where adversarial examples are generated and incorporated as training samples so that the AI models will learn to be robust against them during test time \cite{madry2017towards,zhang2019theoretically,shafahi2019adversarial,zhang2020attacks}. Initially, AT-based defenses came with some limitations such as high computational cost due to the expensive steps of crafting strong adversarial training samples \cite{kannan2018adversarial,xie2019feature}. To reduce the cost of AT, recent works that can reduce the number of optimization steps required for generating adversarial training examples have been proposed \cite{chen2020robust,shafahi2019adversarial,andriushchenko2020understanding}, while maintaining high adversarial accuracy.

Adding to the diversity of defenses, several works improved the robustness of models by minimizing the effects of small perturbations performed on the models' prediction \cite{drucker1991double,ross2018improving,jakubovitz2018improving} or training model to rely on less-superficial features that are more aligned with human vision \cite{etmann2019connection,chan2019jacobian,chan2020thinks}. Another class of defenses is provable defenses which seek to provide a performance guarantee for an AI model's performance in the face of adversarial examples \cite{hein2017formal,raghunathan2018semidefinite,cohen2019certified,balunovic2019adversarial,blum2020random}. Provable defenses can offer assurance to users by providing a theoretically proven worst-case scenario of the AI's performance when under attack. This can help decision-makers weigh the reward-risk trade-off of deploying the AI system with more certainty. 


\textbf{Data Poisoning:}
Modern deep learning-based AI models rely heavily on large amounts of training data, typically mined from public sources such as the Internet or crowdsourcing platforms. This exposes the models to the threat of data poisoning where an attacker can degrade a model's performance by corrupting a small subset of its training data as a data contributor \cite{nelson2008exploiting,xiao2015support, steinhardt2017certified}. The corruption typically involves a combination of altering the data's labels and making edits to the original training samples. A sophisticated variant of data poisoning, called backdoor poisoning, allows an adversary to control a model's prediction through a poison signature in the model's input while eluding detection as the model performs seemingly well on clean inputs \cite{gu2017badnets,Trojannn,adi2018turning,chan2020poison}. 

Several defenses have effectively countered this threat under certain conditions. The first type of defense works by filtering poisoned samples that contain spectral signatures where their hidden states would have different statistics compared to the majority of uncorrupted samples \cite{tran2018spectral}. Initial iterations of such defense either require prior knowledge such as the ratio of poisoned samples \cite{tran2018spectral}, size of the poison patterns \cite{qiao2019defending} or only work for small poison patterns \cite{wangneural,qiao2019defending}. Newer variants are able to overcome these limitations to generalize a wider scope of poisoning scenarios \cite{chen2018detecting,chan2019poison}. Other defense approaches include pruning away `suspicious' neurons that lie dormant in the presence of clean validation data \cite{liu2018fine} or using differential privacy to counter the poison \cite{ma2019data}. More recently, provable defense approaches have been studied in \cite{rosenfeld2020certified,weber2020rab} that can provide guarantees to a model's performance when the information of data poisoning is known.

The defenses and attacks on AI systems will likely see a protracted arms race as long as AI is adopted, especially in high-stake applications. It is, hence, vital to invest resources on the development of more advanced defenses to stay ahead of this race in securing AI models. On top of safety against adversaries, much work has also been studied on improving models' reliability in the face of high signal-to-noise environments \cite{dodge2017study,hendrycks2019natural}. Exposing AI models to training samples augmented with a wide diversity of corruptions has shown to improve the performance of models during such challenging scenarios \cite{cubuk2018autoaugment,hendrycks2020many,kireev2021effectiveness}.

\section{Discussion}

AI is changing our lives in more ways than one would expect, such as in economy, entertainment, games, sociality, healthcare, etc., due to the extremely fast pace of development and adoption of AI technologies. With the powerful capabilities of AI, its functionalities have been extended to almost every single corner, such as the breakthrough of AlphaFold \cite{senior2020improved} in biological domain. It is foreseeable that AI will permeate many other important areas in the near future. However, without the deep consideration of sustainable development of AI, it will violate our original intention of developing AI to create a better world, benefit our society and improve the quality of our lives. We believe now is the right moment to think beyond these huge benefits from AI and invest sufficient efforts in addressing its impacts on social, environmental and resource utilization issues. Fortunately, more and more researchers have realized the critical sustainable development issue when designing AI techniques. In this survey, we have reviewed and summarized AI techniques for the sustainable development on two major perspectives, \textit{i.e.}, environmental and social perspectives.

Carbon footprint of AI is a big concern in \textit{environmental perspective} as AI techniques, especially deep learning, are resource-intensive. From data collection and model training to real-time inference and adaptation, huge amount of energy, memory and human efforts are required. It is clear that AI models are generating huge carbon emissions and consuming intensive computing power, exacerbated by the ever expanding real-world applications being created by technology companies, universities, research institutions, hospitals, factories, government agencies, etc.

To alleviate this critical issue, the main technical challenges are to reduce the sizes of AI models and the amount of training data via emerging techniques in computation-efficient and data-efficient learning. For instance, compressing AI models into smaller size without significantly compromising their performance is a popular way to consume much less energy during their deployment stage. It includes techniques of network pruning, quantization and knowledge distillation, where many advanced works have been developed. For data-efficient learning, the main objective is to reduce/alleviate the use of large labeled dataset for model training, as the collection of large labeled dataset will be time-consuming, utilize intensive human efforts, and lead to huge carbon footprint.

Even though a lot of efforts have been made in addressing the concern of AI models' carbon footprint, it is far from satisfactory in regards to true environmental sustainability of AI. Currently, most of AI research is driven by performance (\textit{e.g.}, accuracy) without taking resource constrains and carbon footprint into consideration. For example, the OpenAI GPT-3 that achieves leading performance on NLP tasks contains 175 billion parameters, which requires approximately 288 years to train on a single powerful V100 GPU \cite{narayanan2021efficient}. Most of institutes/organizations do not have sufficient resources to conduct research on GPT-3. This equipment ``competition'' is totally unsustainable. Pursuing better performance but without considering negative impacts on environments is a dangerous trajectory and will result in wastage of various resources and efforts. This is the right moment to critically think about our evaluation criterion for sustainable development of AI --- what AI models do we prefer or qualified to be used in real-life? 

In addition to the environmental concerns, social impact of AI models is another key issue to be addressed. We have considered social sustainability of AI through the Responsible AI and Rationalizable \& Resilient AI perspectives. For responsible AI, we mainly focused on two vital aspects of fairness and data privacy, which are key ethics. Existing methods on AI fairness try to solve the problem from data perspective, which is not sufficient. Additional emerging techniques to improve model design may help to eventually solve the fairness issue. Data privacy is also a big challenge, as AI models need to be collaboratively trained with data from multiple sources. Federated learning can effectively improve data privacy, but at the expense of additional training and communication costs. These side effects may ruin the benefits created by data privacy techniques.



Another key issue in social sustainability is to achieve rationalizable \& resilient AI for promoting the acceptance of AI systems. Concurrently, most of AI systems are black-box, and users do not know how they make decisions. Such black-box is not acceptable, especially for critical domains like finance, transportation and healthcare. Existing solutions on explainable AI are still in the early stage which can only explain simple models or instances. The true mechanism of advanced models, especially the powerful deep learning models, remains a mystery. The final problem is the resilience of AI models when encountering unknown perturbations, such as adversarial attacks or data poisoning, also known as the safety issue of AI. For example, there are some threats to federated learning via different types of attacks, such as inference attacks, poisoning attacks, and Sybil attacks \cite{lyu2020privacy,Yang2019}.

Another key issue in social sustainability is to achieve rationalizable \& resilient AI so as to promote acceptance of AI systems. Concurrently, most of AI systems are black-boxes, and users do not know how they make decisions. Such black-box approach is not acceptable, especially for critical domains like finance, transportation and healthcare. Existing solutions on explainable AI are still in the early stage of development and they can only explain simple models or instances. The true mechanism of advanced models, especially the powerful deep learning models, remained a mystery. The final problem is the resiliency of AI models when encountering unknown perturbations, such as adversarial attacks or data poisoning, also known as the safety issue of AI. For example, there are some threats to federated learning via different types of attacks, such as inference attacks, poisoning attacks, and Sybil attacks \cite{lyu2020privacy,Yang2019}.









\section{Challenges and Future Research Directions}


To promote further development of AI, its sustainability issue must be well addressed. Various advanced methods have been designed and they have achieved considerable progresses from the technical perspective. However, it is still far from the true or even satisfactory sustainability of AI. In this section, we attempt to discuss and present some potential future research directions in sustainable AI.


Specifically, to achieve \textit{environmental sustainability}, a fundamental way is to improve the intelligence of AI such that it does not require huge training datasets that are time-consuming and labor-intensive to prepare for model training and can easily adapt to various applications. Another way is to change our evaluation metrics to not only pursuing performance in terms of accuracy but also incorporating energy consumption and resource utilization into consideration. In this regard, a safe-sharing community for the trained AI models can be vital in promoting fast development while reducing repeated energy consumption and carbon emissions.


For \textit{social sustainability}, on the other hand, unfolding complex AI models in an understandable perspective can be an efficient strategy to fully explain how AI models process data to yield explainable outcomes. It can also enhance robustness, transparency and reduce bias, as well as protection of data privacy across different AI development stages, including data collection, processing, model design and execution, and evaluation.


\paragraph{Co-designed AI}

Human intelligence is efficient and environmental-friendly, as it can learn from just a few samples and generalize the learned knowledge to different applications. However, existing AI systems are specifically designed for particular applications and are difficult to adapt to other applications. Even though emerging techniques have been developed for different use-cases, such as few-shot learning, transfer learning, self-supervised learning, etc., they can only address problems under a specific assumption, \textit{e.g.}, availability of a large source domain dataset. Besides, single technique may lead to conflicts when contributing to sustainability. For example, an efficient inference model via knowledge distillation may require more training efforts. 

Co-designed AI aims to address the above issues via combining advanced techniques using fusion frameworks. For instance, we can combine self-supervised learning with continuous learning \cite{maltoni2019continuous} to not only achieve sample-efficacy, but also adapt to various applications (\textit{i.e.}, no need to learn multiple models for different yet related applications). Cognitive reasoning \cite{de2011neural} can also be incorporated for better adaptation and learning from past knowledge instead of additional training samples, resulting in both performance improvement and reduction of training efforts.  

\paragraph{Energy-aware Design}

Currently, most of AI systems are solely evaluated by accuracy without considering the costs of resources and human efforts. This leads to larger datasets and bigger AI models, which has significant impacts on environments in terms of carbon footprint and energy consumption. As the natural resources are limited, this crazy development will be forced to come to a stop one day. To achieve sustainable development of AI, we should change our evaluation schemes to consider not only the accuracy but also the consumed resources in data collection, model training and inference. For instance, we should quantify the energy used for training a specific AI model. Then it is possible to jointly optimize energy consumption and the performance of AI models. Besides, the other efforts on data collection and annotation should also be transferred to energy or a new resource utilization criterion for a more comprehensive evaluation. We refer to this framework as energy-aware AI design which will encourage researchers and engineers to pay more attention to the energy and resource consumption at different stages, such as data acquisition, representation learning, model training and deployment, when designing AI systems. 

\paragraph{Model Sharing Community}

Training AI models requires huge datasets and consumes a large amount of energy, especially for some big models like Transformer \cite{vaswani2017attention} and GPT-3 \cite{brown2020language}. In fact, these big and powerful AI models can be reused for various applications due to their universal representation learning capability. Hence, if these models are shared across our AI community, there is no need for everyone to train these big models from scratch, which can save huge energy. Even for different, but related applications, we can adopt the technique of fine-tuning to only train few layers of the trained big models with less data from specific applications. Besides, we can always train a smaller student model for a specific application from the big teacher models by utilizing the big teacher models through the technique of knowledge distillation. Note that the training of the smaller student model with knowledge distillation can be efficient without the training of the big models. Such sharing community will play a key role in saving training efforts of AI models for environmental sustainability and can also promote the innovation and development of AI techniques.



\paragraph{Unfolding Complex AI models}
Existing explainable AI techniques can mainly explain simple models (such as the classification models of K-Nearest Neighbors and Decision Trees, and regression models of Linear Regression) or the relationship between inputs and outputs based on data attribution and feature attribution, which are far too insufficient from the model explainability perspective. With recent AI breakthrough in deep learning, the actual interpretability that we are pursuing is the clear expression on how it works for each single layer of neural networks. CNN Explainer \cite{wangCNNExplainerLearning2020} is a good start in shedding light on deep convolutional neural networks by visualizing all convolutional 
operations. However, it does not explore on how these convolutional operations have been combined to make the final prediction. This issue of model level interpretability is extremely challenging but of great importance. With the fully explainable models, it may also be able to perform self-diagnosis on whether fairness issues exist, which can thoroughly avoid biases of AI models. Besides, the robustness and data privacy aspects can also be enhanced when the operations of AI models are fully understood. 


\paragraph{Preemptive Approach for AI Safety}
Given how wide and diverse the possible AI applications are, it is critical to scrutinize each stage of an AI system's training and deployment processes for possible entry point where a malicious actor can attack. Privileged access management \cite{tep2015taxonomy}, a widely adopted cybersecurity practice, can help to reduce the attack surface (\textit{e.g.}, training data, model weights, etc.) where a malicious attacker has on the AI system. As the AI workflow advances with novel development such as federated learning \cite{yang2019federated}, it is important to also consider and develop defenses for new safety goals such as privacy protection, apart from model performance, that an attacker might seek to undermine. With AI systems deployed in increasingly important applications, 
it is critical that defenses are developed proactively to avoid catastrophic safety incidents before they even happen. 

\section{Conclusion}

In this paper, we have presented a comprehensive technical survey on challenges confronting sustainable AI, with focused discussions on the sustainability of AI through two major perspectives of environmental sustainability and social sustainability of AI. In particular, we first examined computation-efficient and data-efficient AI techniques for addressing environmental sustainability issue. Then, we discussed social sustainability of AI from the angles of responsible AI and rationalizable \& resilient AI. Meanwhile, limitations and challenges, as well as potential research directions in sustainable AI, have been discussed and presented as an effort to emphasize the importance of this topic, to attract researchers' attentions and to rally more research efforts in this critical area. Finally, we believe that AI has great potential to be sustainable and at the same time create huge economic and social impacts in our beautiful blue planet, provided our researchers can continuously factor in environmental, social, resource utilization and performance costs when investigating and improving their critical research problems.


%





\ifCLASSOPTIONcaptionsoff
  \newpage
\fi



%



\bibliographystyle{IEEEtran}
\bibliography{mybib}




\end{document}